\documentclass[conference]{IEEEtran}
\IEEEoverridecommandlockouts
\usepackage{amsmath}  

\newcommand{\ignore}[1]{}

\newcommand{\ba}{\begin{array}}
        \newcommand{\ea}{\end{array}}
\newcommand{\bc}{\begin{center}}
        \newcommand{\ec}{\end{center}}
\newcommand{\be}{\begin{enumerate}}
        \newcommand{\ee}{\end{enumerate}}
\newcommand{\bea}{\begin{eqnarray}}
        \newcommand{\eea}{\end{eqnarray}}
\newcommand{\beas}{\begin{eqnarray*}}
        \newcommand{\eeas}{\end{eqnarray*}}
\newcommand{\beq}{\begin{equation}}
        \newcommand{\eeq}{\end{equation}}
\newcommand{\bfig}{\begin{figure}}
        \newcommand{\efig}{\end{figure}}
\newcommand{\bi}{\begin{itemize}}
        \newcommand{\ei}{\end{itemize}}
\newcommand{\bpic}{\begin{picture}}
        \newcommand{\epic}{\end{picture}}
\newcommand{\btabular}{\begin{tabular}}
        \newcommand{\etabular}{\end{tabular}}
\newcommand{\btable}{\begin{table}}
        \newcommand{\etable}{\end{table}}

\newcommand{\es}{\vfill
    \rule[-6mm]{170mm}{0.7mm} \\
    \redw{{\tiny
                \hfill S-\theslide}}
    \end{slide}}

\newcommand{\matxx}[1]{{\mathbf #1}}
\newcommand{\vecXX}[1]{{\mathbf {#1}}}
\newcommand{\vecYY}[1]{{\boldsymbol {#1}}}

\newcommand{\argmin}{\operatornamewithlimits{arg\ min}}

\def \hbar {{\bar{h}}}

\def \vecm {{\vecXX{m}}}

\def \vecq {{\vecXX{q}}}

\def \vect {{\vecXX{t}}}

\def \matC {{\matxx{C}}}
\def \matD {{\matxx{D}}}

\def \matQ {{\matxx{Q}}}
\def \matR {{\matxx{R}}}

\def \matT {{\matxx{T}}}

\def \matX {{\matxx{X}}}

\newcommand{\mattwo}[4]{\left[\begin{array}{cc}#1&#2\\#3&#4\end{array}\right]}

\makeatletter
\renewcommand*\env@matrix[1][*\c@MaxMatrixCols c]{%
    \hskip -\arraycolsep
    \let\@ifnextchar\new@ifnextchar
    \array{#1}}
\makeatother

\newcommand{\RR}{\mathbb{R}}

\newcommand{\SO}[1]{\ensuremath{\mathbf{SO}(#1)}}

\newcommand{\Sim}[1]{\ensuremath{\mathbf{Sim}(#1)}}

\font\Bigmath=cmsy10 scaled \magstep2
\def\dplus{\mathrel{%
        \ooalign{$+$\cr\hss\lower.255ex\hbox{\Bigmath\char5}\hss}}}
\def\dminus{\mathrel{%
        \ooalign{$-$\cr\hss\lower.255ex\hbox{\Bigmath\char5}\hss}}}

\newcommand{\cF}{\mathcal{F}}

\newcommand{\cM}{\mathcal{M}}

\def\bea#1\eea{\begin{align}#1\end{align}}
\def\beas#1\eeas{\begin{align*}#1\end{align*}}

\usepackage{xspace}

\usepackage{multirow} 
\usepackage{boldline}

\newcommand{\Fmaster}{\cF_{M}}

\newcommand{\mbf}[1]{\mathbf{#1}}
\newcommand{\mcal}[1]{\mathcal{#1}}

\newcommand{\normsq}[1]{\left\Vert#1\right\Vert^2}
\newcommand{\ray}[1]{\psi\left(#1\right)}

\newcommand{\match}[2]{\vecm_{#1,#2}}
\newcommand{\matchconf}[2]{\vecq_{#1,#2}}

\newcommand{\I}{\mcal{I}}
\newcommand{\Ii}{\I^i}
\newcommand{\Ij}{\I^j}
\newcommand{\Ik}{\I^k}
\newcommand{\If}{\I^f}

\newcommand{\pix}{\mbf{p}}

\newcommand{\X}{\matX}
\newcommand{\Xii}{\X^i_{i}}
\newcommand{\Xji}{\X^j_{i}}

\newcommand{\Xkf}{\X^k_f}

\newcommand{\Xff}{\X^f_f}

\newcommand{\Xffm}{\X^f_{f, m}}

\newcommand{\Xcanon}{\tilde{\matX}}

\newcommand{\Xcanonkk}{\Xcanon^k_{k}}
\newcommand{\Xcanonkkn}{\Xcanon^k_{k,n}}
\newcommand{\Xcanoniim}{\Xcanon^i_{i,m}}
\newcommand{\Xcanonjjn}{\Xcanon^j_{j,n}}

\newcommand{\Ccanon}{\tilde{\matC}}
\newcommand{\Ccanonkk}{\Ccanon^k_{k}}

\newcommand{\C}{\matC}
\newcommand{\Cii}{\C^i_{i}}
\newcommand{\Cji}{\C^j_{i}}

\newcommand{\Ckf}{\C^k_f}

\newcommand{\Q}{\matQ}
\newcommand{\Qii}{\Q^i_{i}}
\newcommand{\Qji}{\Q^j_{i}}

\newcommand{\D}{\matD}
\newcommand{\Dii}{\D^i_{i}}
\newcommand{\Dji}{\D^j_{i}}

\newcommand{\Tkf}{\matT_{kf}}

\newcommand{\Tij}{\matT_{ij}}

\newcommand{\vectau}{\vecYY{\tau}}
\newcommand{\KF}{\mathcal{K}}

\usepackage{cite}
\usepackage{amsmath,amssymb,amsfonts}
\usepackage{algorithmic}
\usepackage{graphicx}
\usepackage{textcomp}
\usepackage{xcolor}
\newcommand{\redx}{{\color{red}\texttimes}}
\usepackage{booktabs}
\def\BibTeX{{\rm B\kern-.05em{\sc i\kern-.025em b}\kern-.08em
    T\kern-.1667em\lower.7ex\hbox{E}\kern-.125emX}}
\begin{document}

\title{Multi-Agent Monocular Dense SLAM  \\ With 3D Reconstruction Priors
}

\author{

\IEEEauthorblockN{Yuchen Zhou}
\IEEEauthorblockA{\textit{Xi'an Jiaotong-Liverpool University} \\
Suzhou, China  \\
Email: 1796382054@qq.com}

\and

\IEEEauthorblockN{Haihang Wu*}
\IEEEauthorblockA{
\textit{ Southeast University} \\
Nanjing, China \\
Email: hthh1211@sina.com}

}

\maketitle

\begin{abstract}
Monocular Simultaneous Localization and Mapping (SLAM) aims to estimate a robot's pose while simultaneously reconstructing an unknown 3D scene using a single camera. While existing monocular SLAM systems generate detailed 3D geometry through dense scene representations, they are computationally expensive due to the need for iterative optimization. To address this challenge, MASt3R-SLAM utilizes learned 3D reconstruction priors, enabling more efficient and accurate estimation of both 3D structures and camera poses. However, MASt3R-SLAM is limited to single-agent operation. In this paper, we extend MASt3R-SLAM to introduce the first multi-agent monocular dense SLAM system. Each agent performs local SLAM using a 3D reconstruction prior, and their individual maps are fused into a globally consistent map through a loop-closure-based map fusion mechanism. Our approach improves computational efficiency compared to state-of-the-art methods, while maintaining similar mapping accuracy when evaluated on real-world datasets.

\end{abstract}


\section{Introduction}
\subsection{Single-Agent SLAM}
Visual Simultaneous Localization and Mapping (vSLAM) aims to estimate a mobile robot’s trajectory while reconstructing the 3D structure of an unknown environment using only camera input. \cite{Davison2007MonoSLAM:SLAM}. vSLAM has found widespread application in robotic navigation and augmented reality, thanks to its ability to provide spatial awareness.

Monocular SLAM, a subset of vSLAM, uses only a camera and has gained significant attention due to its simple configuration and low cost. Early monocular SLAM relied on sparse feature-based scene representations to reduce computational and memory demands~\cite{Mur-Artal2015ORB-SLAM:System}. Although such approaches enable accurate camera tracking with limited resources~\cite{Campos2021ORB-SLAM3:SLAM}, the resulting sparse maps lack geometric detail and are therefore unsuitable for tasks that require dense spatial understanding, such as robot navigation. With the advent of high-performance hardware such as GPUs, research has shifted toward dense representations—such as point clouds~\cite{Newcombe2011DTAM:Real-Time,Engel2014LSD-SLAM:SLAM} and Truncated Signed Distance Functions (TSDFs)~\cite{Newcombe2011KinectFusion:Tracking}. These representations capture detailed geometry but struggle with view-dependent appearance and suffer from high memory consumption in large scenes.

To overcome this limitation, existing work has explored continuous, joint geometry–appearance representations. Neural Radiance Fields (NeRF) model the scene as a continuous volumetric function parameterized by a neural network, enabling photorealistic rendering and novel view synthesis. Despite their compactness and expressiveness, NeRF-based SLAM systems remain computationally expensive~\cite{Sucar2021IMAP:Real-Time,Zhu2022NICE-SLAM:SLAM}, since the network must be evaluated hundreds of times per pixel during image rendering. Recent advances in 3D Gaussian splatting alleviate this problem by representing the scene as a set of Gaussian primitives, achieving comparable visual fidelity with substantially higher efficiency~\cite{Matsuki2024GaussianSLAM,Sandstrom2025Splat-SLAM:Gaussians}. Nonetheless, these approaches still require iterative optimization for state estimation, including camera pose and scene reconstruction.

To accelerate state estimation, studies have trained neural networks on large 3D datasets. These models can predict camera poses and scene structure in a single forward pass, significantly improving efficiency. DUST3R~\cite{Wang2024DUSt3R:Easy} introduces the pointmap representation to model 3D shape and trains a transformer to predict it from image pairs. MASt3R~\cite{Leroy2024GroundingMASt3R} improves image matching accuracy and efficiency by adding a network head that outputs dense local features. Subsequent work extends these ideas to handle hundreds of input views, enabling consistent multi-view 3D reconstruction~\cite{Wang2025VGGT:Transformer}. However, these approaches are primarily designed for Structure-from-Motion (SfM), which reconstructs camera trajectories and 3D scene structure from a set of pre-captured images \textit{offline}. 

Recent studies have adapted MASt3R for online and incremental SLAM. Building on MASt3R, MASt3R-SLAM~\cite{Murai2025MASt3R-SLAM:Priors} estimates pointmaps from image pairs and fuses them for local tracking (camera pose estimation) and mapping (3D structure reconstruction). Loop closures are employed to enhance pose accuracy and map consistency. Despite its strong performance, MASt3R-SLAM remains a single-agent system. While single-agent systems are sufficient for tasks like autonomous navigation in small environments, they are not enough for many applications, such as search and rescue or warehouse automation. In these scenarios, multiple robots must collaborate for efficient exploration, task execution, and data sharing, highlighting the need for multi-agent SLAM.

\subsection{Multi-Agent SLAM}

Existing multi-agent SLAM systems generally adopt either a distributed or centralized architecture. In distributed approaches, each agent estimates its pose and constructs a local map independently, exchanging information only intermittently—an arrangement suitable for environments with limited communication capabilities~\cite{Lajoie2024Swarm-SLAMSystems}. Centralized systems, in contrast, depend on a central server that collects local maps and trajectories from all agents to perform global optimization~\cite{Yugay2025MAGiC-SLAM:SLAM,Campos2021ORB-SLAM3:SLAMb}. In this study, we adopt a centralized system due to its superior 3D reconstruction quality

Significant progress has been made in multi-agent SLAM. Most systems employing sensor configurations like monocular+IMU~\cite{Schmuck2021COVINS:Collaboration}, stereo~\cite{Cieslewski2018Data-EfficientSLAM}, or stereo+IMU~\cite{Lajoie2020DOOR-SLAM:Teams, Xu2024D2SLAM:Swarm, Tian2021Kimera-Multi:Systems, Lajoie2024Swarm-SLAMSystems}. While monocular cameras are more affordable and simpler, their use in multi-agent SLAM remains limited, primarily due to challenges such as arbitrary map scale and low feature density. To the best of our knowledge, CCM-SLAM~\cite{Schmuck2018CCM-SLAM:Teams} is one of the few multi-agent systems utilizing monocular cameras. However, it relies on sparse features for localization and mapping, which constrains its applicability in tasks that require dense spatial awareness.

In this work, we extend MASt3R-SLAM~\cite{Murai2025MASt3R-SLAM:Priors} and propose the first multi-agent monocular dense SLAM framework. Each agent is equipped with MASt3R, a strong 3D reconstruction prior that enables accurate tracking and mapping. The resulting local trajectories and maps are then integrated into a consistent global map through a loop-closure-based fusion mechanism. The main contributions of this study are as follows:

\begin{enumerate}
\item We introduce the first multi-agent dense SLAM system leveraging 3D reconstruction priors.
\vspace{0.2cm}
\item We present a loop-closure-based local map fusion method that generates globally consistent maps.
\vspace{0.2cm}
\item We demonstrate that our system achieves competitive accuracy while significantly outperforming state-of-the-art methods in terms of computational speed.
\vspace{0.2cm}
\end{enumerate}

\section{Method}
\begin{figure*}[ht]
\centering
\includegraphics[width=\linewidth]{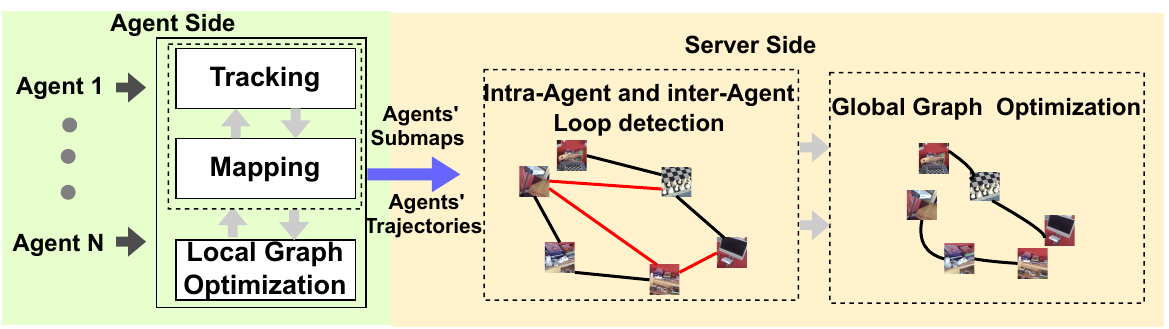}
\caption{System overview. Agent Side: Each agent processes its own RGBD stream, performing tasks including tracking, mapping, and local graph optimization. Upon completion, the agent sends its sub-map and trajectory to the central server. Server Side: The server performs both intra-agent and inter-agent loop closure detection, followed by global graph optimization. This process merges the submaps into a unified global map and updates the agent poses accordingly.}
\label{fig:system_diagram}
\end{figure*}

Our framework is illustrated in Figure~\ref{fig:system_diagram}. Each agent is equipped with an identical pre-trained 3D vision model, MASt3R (Section~\ref{background}). Upon receiving an RGB image stream, each agent independently performs local tracking, mapping, and loop closure (Section~\ref{distributed slam}) based on MASt3R’s learned 3D priors. After all agents complete their local SLAM processes, their submaps and trajectories are transmitted to a central server, which fuses them into a globally consistent map (Section~\ref{global map}).

\subsection{Background}
\label{background}
In this section, we revisit MASt3R. MASt3R is a pre-trained model that reconstructs the 3D structure from two-view input:

\begin{equation}
(\Xii, \Xji, \Cii, \Cji, \Dii, \Dji, \Qii, \Qji)
= \Fmaster(\Ii, \Ij),
\end{equation}
where $\Ii, \Ij \in \RR^{H\times W\times3}$ are input images, 
$\Xii, \Xji \in \RR^{H\times W\times3}$ are the corresponding pointmaps represented in the coordinate frame of camera $i$, 
$\Cii, \Cji \in \RR^{H\times W\times1}$ are their confidences, 
$\Dii, \Dji \in \RR^{H\times W\times d}$ are $d$-dimensional matching features, 
and $\Qii, \Qji \in \RR^{H\times W\times1}$ are their confidences.

To enable tracking and mapping in SLAM,  pixel correspondences, $\match{i}{j} = \cM(\X^i_{\smash{i}}, \X^j_{\smash{i}}, \D^i_{\smash{i}}, \D^j_{\smash{i}})$, between image pairs should be determined. 
MASt3R SLAM achieves this efficiently by converting the pointmaps $\X$ into unit-norm rays $\ray{\X}$. 
For each 3D point $\mbf{x} \in \Xji$, the associated pixel $\pix$ in the reference frame is obtained by minimizing the ray discrepancy
\beq
    \pix^* = \argmin_{\pix} \normsq{\ray{[\Xii]_\pix} - \ray{\mbf{x}}},
    \label{eq:ray_p}
\eeq
which equivalently minimizes the angle $\theta$ between rays:
\beq
    \normsq{\psi_1 - \psi_2} = 2(1-\cos{\theta}), \quad \cos{\theta} = \psi_1^T \psi_2.
    \label{eq:ray_error}
\eeq
This nonlinear least-squares problem is solved via the Levenberg-Marquardt algorithm, producing initial pixel correspondences $\match{i}{j}$. 
To refine these matches, a coarse-to-fine search is performed in image space using the per-pixel features $\Dii$ and $\Dji$. Once the pixel correspondences between two images are obtained, their relative camera pose can be estimated (Section~\ref{distributed slam}). We represent each pose as $\matT \in \Sim{3}$, defined by:
\beq
\matT = \mattwo{s\matR}{\vect}{0}{1}~, \quad
\eeq
where $\matR \in \SO{3}$ is the  rotation matrix, $\vect \in \RR^3$ is the translation vector, and  $s \in \RR$ is the scale.

\subsection{Local SLAM on the Agent Side}
\label{distributed slam}
In this section, we extend MASt3R-SLAM from a single-agent to a multi-agent setting. Each agent operates on an independent GPU with dedicated CPU cores and memory. Agents process distinct RGB streams and independently perform local tracking, mapping, and loop closure using MASt3R, without any inter-agent communication.

\textbf{Initialization:} at the beginning of the SLAM pipeline, each agent performs monocular inference using the pre-trained MASt3R model to obtain the initial 3D pointmap and its confidence. This first frame, along with its pointmap, is stored as the initial keyframe for subsequent tracking.

\textbf{Tracking:} for tracking, the last keyframe $\Ik$ with its canonical pointmap $\Xcanonkk$ is available. 
For each incoming frame $\If$, we feed $(\If, \Ik)$ into MASt3R to obtain the current frame's pointmap $\Xff$. 
Pixel correspondences $\match{f}{k}$ between $\If$ and $\Ik$ are obtained as described in Section~\ref{background}. 
For each matched pixel, $\Xff$ is transformed by $\Tkf$ to align with the canonical keyframe point $\Xcanonkk$, and the relative pose $\Tkf$ is estimated by minimizing the ray error~\cite{Murai2025MASt3R-SLAM:Priors}:

\begin{equation}
E_r = \sum_{{m, n} \in \match{f}{k}} 
\left\| \frac{\ray{\Xcanonkkn} - \ray{\Tkf \Xffm}}{w(\matchconf{m}{n}, \sigma^2_r)} \right\|_{\rho},
\end{equation}

where $\matchconf{m}{n}$ weights correspondences by confidence, and the Huber norm $\|\cdot\|_\rho$ and per-match weight  $w$ improve robustness against outliers. The relative pose $\Tkf$ is iteratively refined using Gauss-Newton in an IRLS framework~\cite{Murai2025MASt3R-SLAM:Priors}. Let $\vectau \in \mathfrak{sim}(3)$ denote a small perturbation to the current estimate of $\Tkf$. 
The update rule is:

\begin{equation}
\Tkf \leftarrow \vectau \oplus \Tkf.
\end{equation}

\textbf{Mapping:} After estimating $\Tkf$, the canonical keyframe pointmap $\Xcanonkk$ is incrementally updated using a confidence-weighted average approach~\cite{Murai2025MASt3R-SLAM:Priors}:

\begin{equation}
\Xcanonkk \leftarrow \frac{\Ccanonkk \Xcanonkk + \Ckf (\Tkf \Xkf)}{\Ccanonkk + \Ckf}, \quad
\Ccanonkk \leftarrow \Ccanonkk + \Ckf,
\end{equation}

This procedure fuses new observations into the canonical keyframe without storing all previous predictions, improving mapping accuracy and robustness while remaining computationally efficient.

\textbf{New Keyframe Insertion:} During tracking, the system determines whether the current frame should become a new keyframe by evaluating the fraction of valid matches and the number of pixels matched with the previous keyframe. If either metric falls below a predefined threshold, a new keyframe $\KF_i$ is instantiated

\textbf{Local Graph Optimization:} The tracking and mapping procedures described above provide only local state estimation between consecutive keyframes, causing estimation errors to accumulate over time. To mitigate this drift, we incrementally construct a factor graph that connects the current keyframe to relevant past keyframes and jointly optimize their poses. Specifically, each new keyframe is first linked to its immediate predecessor to maintain temporal consistency. The system then queries a retrieval database using the current keyframe’s encoded features to identify visually similar keyframes from earlier in the sequence. For each retrieved match, an edge is added to the factor graph, forming additional spatial constraints. After establishing all connections, the system performs Gauss–Newton optimization on the factor graph to minimize ray-based reprojection errors, refining keyframe poses and enhancing the agent's submap consistency.

\subsection{ Global Map Fusion on the Server Side}
\label{global map}
After all agents complete their local processing, their submaps are fused into a globally consistent map. All keyframes from each agent are transferred to a global keyframe buffer. For each keyframe, the system queries a retrieval database to identify visually similar keyframes both within the same agent (intra-agent loops) and across different agents (inter-agent loops). To preserve temporal continuity, each keyframe is also connected to its immediately preceding keyframe. These connections form geometric constraints in a global factor graph, which is then jointly optimized using the Gauss–Newton method to minimize ray-based residuals and refine all keyframe poses:
\begin{equation}
E_g {=}\sum_{{i,j} \in \mathcal{E}}\sum_{{m, n} \in \match{i}{j}}
\left\|
\frac{
\ray{\Xcanoniim} - \ray{\Tij \Xcanonjjn}
}{
w(\matchconf{m}{n}, \sigma^2_r)
}
\right\|_{\rho},
\label{eq:backend_ray}
\end{equation}
where $\mathcal{E}$ is the set of edges in the global factor graph, and $\Tij$ is the relative transformation from keyframe $j$ to $i$.

\section{Results}

\textbf{Baselines:} To evaluate tracking, we compare our method with state-of-the-art multi-agent SLAM systems, including SWARM-SLAM~\cite{Lajoie2024Swarm-SLAMSystems}, CCM-SLAM~\cite{Schmuck2018CCM-SLAM:Teams}, CP-SLAM~\cite{Hu2023CP-SLAM:System}, and MAGiC-SLAM~\cite{Yugay2025MAGiC-SLAM:SLAM}. Only CCM-SLAM is a monocular SLAM system, while the other baselines are RGB-D SLAM systems that incorporate depth maps for tracking and mapping. 

\textbf{Datasets:} We test our approach on both simulated dataset Multiagent Replica~\cite{Hu2023CP-SLAM:System} and real-world dataset Aria datasets~\cite{Pan2023AriaPerception}. Three two-agent RGB-D sequences in the  Multiagent Replica benchmark are used for evaluation; two sequences contain 2,500 frames each, while the Office-0 sequence includes 1,500 frames. For real-world evaluation, we select three extended, predominantly static sequences from two rooms in the Aria dataset, and extract 500 consecutive frames from each for evaluation~\cite{Yugay2025MAGiC-SLAM:SLAM}.

\begin{table}[ht]
    \caption{Tracking performance on Aria dataset
    (ATE RMSE [cm]$\downarrow$). \redx{} denotes invalid results caused by CCM-SLAM failures, while \textbf{--} indicates that CP-SLAM does not support configurations with more than two agents.}
    \centering
    \resizebox{\columnwidth}{!}{%
    \begin{tabular}{@{}llcc@{}}
    
    \toprule
    \textbf{Method} & \textbf{Agent} & \textbf{Room0} & \textbf{Room1} \\
    \midrule
    CCM-SLAM~\cite{Schmuck2018CCM-SLAM:Teams} &    \textbf{Agent 1}      & \redx & \redx \\
    Swarm-SLAM~\cite{Lajoie2024Swarm-SLAMSystems} &       & 6.11 & 4.29 \\
    CP-SLAM~\cite{Hu2023CP-SLAM:System} &          & 0.68 & 5.06 \\
    MAGiC-SLAM~\cite{Yugay2025MAGiC-SLAM:SLAM} &     & 0.67 & 0.96 \\ 
    Ours (Uncalibrated) & & 8.01 & 18.48 \\
    Ours (Calibrated) & & 1.63 & 1.50\\
    \midrule
    CCM-SLAM~\cite{Schmuck2018CCM-SLAM:Teams} &   \textbf{Agent 2}         & \redx & \redx \\
    Swarm-SLAM~\cite{Lajoie2024Swarm-SLAMSystems} &         & 8.43 & 4.95 \\
    CP-SLAM~\cite{Hu2023CP-SLAM:System} &            & 5.39 & 0.68 \\
    MAGiC-SLAM~\cite{Yugay2025MAGiC-SLAM:SLAM} &     & 1.13 & 0.53 \\    
     Ours (Uncalibrated) & &7.25 & 61.34 \\
    Ours (Calibrated) & & 1.16 & 1.08\\
    \midrule
    CCM-SLAM~\cite{Schmuck2018CCM-SLAM:Teams} &   \textbf{Agent 3}        & \redx & \redx \\
    Swarm-SLAM~\cite{Lajoie2024Swarm-SLAMSystems} &        & 4.82 & 5.12 \\
    CP-SLAM~\cite{Hu2023CP-SLAM:System} &           & \textbf{--} & \textbf{--} \\
    MAGiC-SLAM~\cite{Yugay2025MAGiC-SLAM:SLAM} &    & 1.67 & 0.46 \\
    Ours (Uncalibrated) & &  7.51 & 13.63\\
    Ours (Calibrated) & & 0.51 & 0.45\\
    \midrule
    CCM-SLAM~\cite{Schmuck2018CCM-SLAM:Teams} &  \textbf{Average}                & \redx & \redx \\
    Swarm-SLAM~\cite{Lajoie2024Swarm-SLAMSystems} &               & 6.45 & 4.78 \\
    CP-SLAM~\cite{Hu2023CP-SLAM:System} &                  & 3.03 & 2.87 \\
    MAGiC-SLAM~\cite{Yugay2025MAGiC-SLAM:SLAM} &           & 1.15 & 0.65 \\
    Ours (Uncalibrated) & & 7.59  & 31.15\\
    Ours (Calibrated) & &  1.10 & 1.01 \\
    \bottomrule
    \end{tabular}%
    }
    \label{tab:tracking_aria_multiagent}
\end{table}

\begin{table}[ht]
    \caption{Tracking performance on Multiagent Replicadataset
    (ATE RMSE [cm]$\downarrow$).  \redx{} denotes invalid results caused by CCM-SLAM failures.}
    \centering
    \resizebox{\columnwidth}{!}{%
    \begin{tabular}{@{}llccc@{}}
    \toprule
    \textbf{Method} & \textbf{Agent} & \textbf{Off-0} & \textbf{Apt-0} & \textbf{Apt-1} \\
    \midrule
    CCM-SLAM~\cite{Schmuck2018CCM-SLAM:Teams} &       \textbf{Agent 1}   & 9.84 & \redx & 2.12 \\
    Swarm-SLAM~\cite{Lajoie2024Swarm-SLAMSystems} &               & 1.07 & 1.61 & 4.62  \\
    CP-SLAM~\cite{Hu2023CP-SLAM:System} &                  & 0.50 & 0.62 & 1.11  \\
    MAGiC-SLAM~\cite{Yugay2025MAGiC-SLAM:SLAM} &                    & 0.31 & 0.13 & 0.21 \\  
    Ours (Calibrated) & & 3.79 &3.27 & 4.60  \\
    \midrule
    CCM-SLAM~\cite{Schmuck2018CCM-SLAM:Teams} &        \textbf{Agent 2}  & 0.76 & \redx & 9.31\\
    Swarm-SLAM~\cite{Lajoie2024Swarm-SLAMSystems} &        & 1.76 & 1.98 & 6.50 \\
    CP-SLAM~\cite{Hu2023CP-SLAM:System} &           & 0.79 & 1.28 & 1.72  \\
    MAGiC-SLAM~\cite{Yugay2025MAGiC-SLAM:SLAM} &             & 0.24 & 0.21 & 0.30  \\   
     Ours (Calibrated)  &  &1.34 & 7.20& 6.35 \\
    \midrule
    CCM-SLAM~\cite{Schmuck2018CCM-SLAM:Teams} &      \textbf{Average} & 5.30 & \redx & 5.71  \\
    Swarm-SLAM~\cite{Lajoie2024Swarm-SLAMSystems} &     & 1.42 & 1.80 & 5.56  \\
    CP-SLAM~\cite{Hu2023CP-SLAM:System} &        & 0.65 & 0.95 & 1.42  \\
    MAGiC-SLAM~\cite{Yugay2025MAGiC-SLAM:SLAM} &          & 0.27 & 0.16 & 0.26  \\
     Ours (Calibrated)  & & 2.57  &5.24 & 5.48 \\
    \bottomrule
    \end{tabular}%
    }
    \label{tab:tracking_replica_multiagent}
\end{table}

\subsection{Camera Pose Estimation}
The RMSE of absolute trajectory error (ATE [CM]) is reported to assess the tracking performance.

\textbf{MultiagentAria:} We evaluate our method on the real-world Aria dataset. The data presented in Table~\ref{tab:tracking_aria_multiagent} demonstrate that our SLAM system achieves tracking accuracy on par with the state-of-the-art method MAGiC-SLAM, while substantially surpassing other baseline approaches. Additionally, we show that our method is much faster than MAGiC-SLAM (Section~\ref{Ablation}). Notably, MAGiC-SLAM is an RGB-D SLAM system that leverages ground truth depth maps for camera tracking, while our method relies exclusively on RGB images for tracking. We also observe that the prior on camera intrinsics significantly impacts performance.  We report the performance of  CP-SLAM using only the first two agents in our setup,  as it does not support multi-agent operation beyond two agents.

\textbf{MultiagentReplica:} We also evaluate our method against baselines on the synthetic MultiagentReplica dataset. Our approach performs similarly to CCM-SLAM, both of which are RGB-based SLAM systems, but lags behind the RGB-D SLAM baselines. This performance gap is primarily due to the pre-trained MASt3R model, which was optimized for a standard input resolution of 512×512. Consequently, the original image resolution of 1200×680 in the MultiagentReplica dataset had to be downscaled to 512×512, resulting in the loss of important visual information and potential distortion. In contrast, the Aria dataset consists of images already at the 512×512 resolution, preserving more visual detail and leading to improved performance.

\begin{figure*}[ht]
\centering
\includegraphics[width=\linewidth]{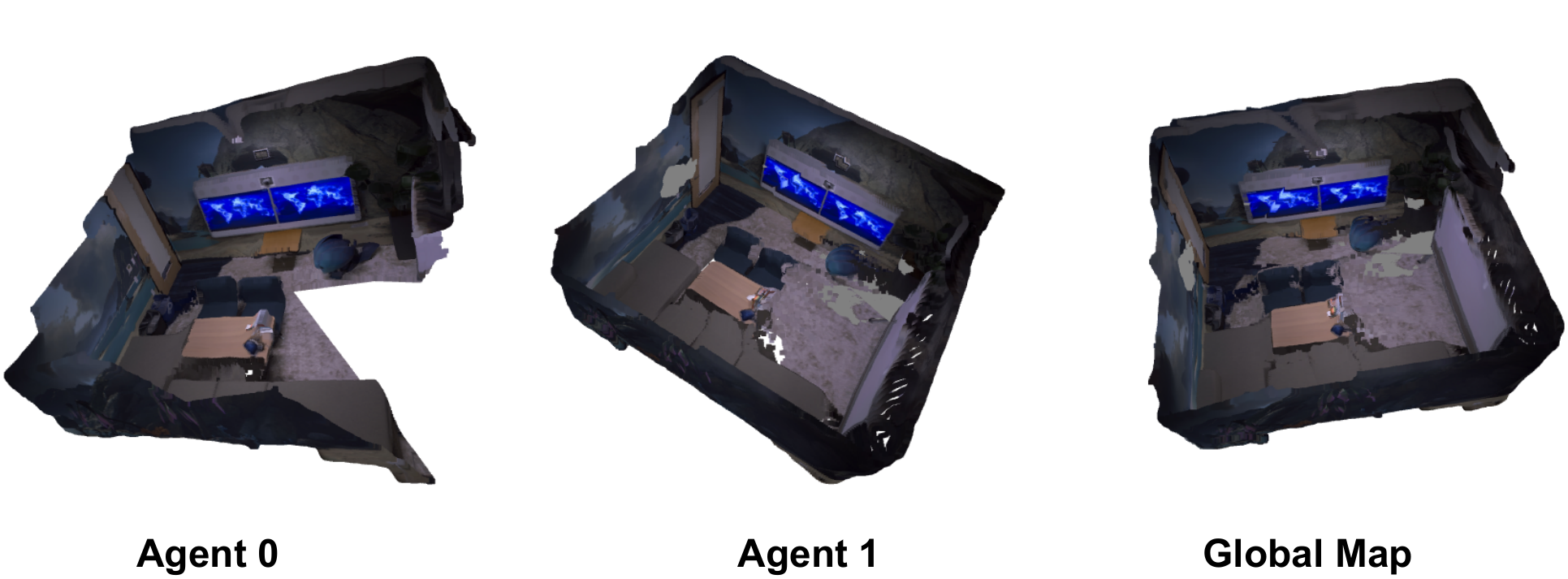}
\caption{Reconstruction of Office 0 in the Replica dataset, where each agent independently reconstructs the scene. The individual maps are then merged into a global map using our global map fusion algorithm. }
\label{fig:office0_reconstruction}
\end{figure*}

\begin{figure*}[ht]
\centering
\includegraphics[width=\linewidth]{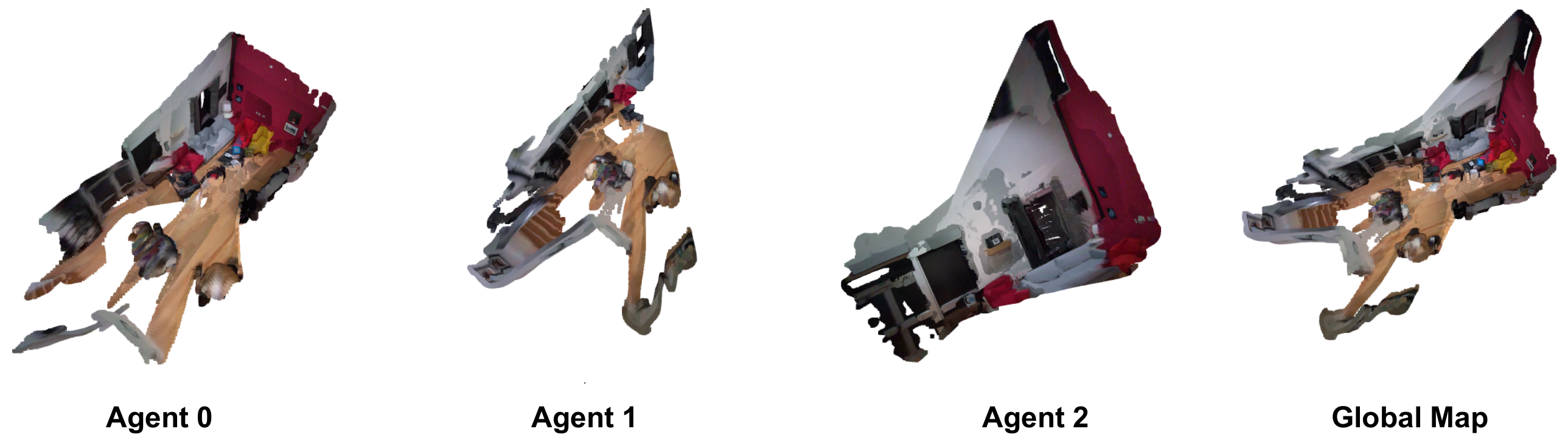}
\caption{Reconstruction of Room 0 in the Aria dataset, where each agent independently reconstructs the scene, and their maps are subsequently fused into a global map using our global map fusion algorithm.. }
\label{fig:room0_reconstruction}
\end{figure*}

\subsection{Dense Geometry Evaluation}
For 3d scene reconstruction, our method is not directly comparable to existing baselines, as it represents the first distributed dense monocular SLAM system. While most baselines rely on depth information and represent the 3D scene using Gaussian primitives, our approach leverages monocular cameras to produce dense point clouds. CCM-SLAM is the only exception, utilizing monocular cameras; however, it generates sparse feature maps, whereas our method yields dense point clouds. Given the differences in scene representation and the use of depth information, a direct and fair comparison of 3D reconstruction quality is not feasible. As such, we report only the 3D reconstruction quality of our method.

To establish the ground truth for evaluation, we first create a reference point cloud by back-projecting the ground truth depth images using the dataset's camera poses~\cite{Murai2025MASt3R-SLAM:Priors}. We then align this reference point cloud to the estimated point cloud using ICP.

\begin{table}[t]\centering
\caption{Reconstruction Evaluation on MultiagentAria and MultiagentReplica with all metrics in metres.}\label{tab:geometry}
\scriptsize
\begin{tabular}{lcccc}
 \toprule
\textbf{MultiagentAria} & \textbf{Agent} &Accuracy &Completion &Chamfer \\
\hline
\multirow{3}{*}{Room 0} & Agent 1 & 0.0478 & 0.1779 & 0.1129\\
  & Agent 2 &0.1006 &0.1671 & 0.1339 \\
    & Agent 3 &0.3690 & 0.4170&   0.3930\\
\multirow{3}{*}{Room 1} & Agent 1 & 0.0574 & 0.0619& 0.0597\\
  & Agent 2 &0.0811 & 0.0885 &  0.0848\\
    & Agent 3 & 0.1137 & 0.0352 & 0.0744 \\
    & Average & 0.1283
 & 0.1579
 & 0.1431
 \\
\hline
\textbf{MultiagentReplica} & \textbf{Agent} & Accuracy &Completion &Chamfer \\
\hline
\multirow{2}{*}{Office 0} & Agent 1  & 0.0344 &   0.0655 & 0.0500 \\
  & Agent 2 & 0.0115 & 0.0975 &  0.0545\\
\multirow{2}{*}{Apartment 0} & Agent 1  & 0.1793 & 0.2671 & 0.2232\\
  & Agent 2 & 0.0305 & 0.0763 &  0.0534\\
\multirow{2}{*}{Apartment 1} & Agent 1  & 0.0154& 0.0636& 0.0395\\
  & Agent 2 & 0.0142 &0.0440 &   0.0291\\
\multirow{2}{*}{Apartment 2} & Agent 1  & 0.0403& 0.0792 & 0.0597\\
  & Agent 2 & 0.0287 & 0.0758 & 0.0522 \\
    & Average & 0.0443
 & 0.0961
 &  0.0702\\
\bottomrule
\end{tabular}
\label{tab:dense geometry evaluation}
\end{table}

We evaluate performance using three metrics: Accuracy, Completion, and Chamfer Distance. Accuracy is defined as the root mean square error (RMSE) between each estimated point and its nearest ground truth point, while Completion measures the distance from each reference point to its closest estimated point. Both metrics are calculated within a 0.5 m threshold. The final Chamfer Distance is reported as the mean of Accuracy and Completion.

We summarize the results in Table~\ref{tab:dense geometry evaluation}. Our method demonstrates reasonable reconstruction performance, with average Chamfer Distances of 0.1431 and 0.0702 for the Multiagent-Aria and Multiagent-Replica datasets, respectively. These results are comparable with the single-agent SLAM system, MAST3R-SLAM~\cite{Murai2025MASt3R-SLAM:Priors}, which reports average Chamfer Distances of 0.056 and 0.09 on the 7-Scenes~\cite{Shotton2013SceneImages} and EuRoC~\cite{Burri2016TheDatasets} datasets, respectively. This underscores the significance of 3D vision priors in enhancing 3D scene reconstruction. Furthermore, graph optimization plays a crucial role in improving reconstruction quality.

\subsection{Qualitative Results}

We present examples of dense reconstructions for the Replica dataset in Figure~\ref{fig:office0_reconstruction} and for the Aria dataset in Figure~\ref{fig:room0_reconstruction}. These figures demonstrate that individual agents are capable of reconstructing the scene effectively from their local observations. Leveraging our global map fusion module, these agent-specific maps can be seamlessly integrated into a unified global map with a reasonable reconstruction quality.

\subsection{Ablation Study}
\label{Ablation}
\textbf{Runtime Analysis on  ReplicaMultiagent
 office0:} We compare the computational speed of our method with the state-of-the-art MAGiC-SLAM, using NVIDIA A100 GPUs. Our method achieves an overall speed of 11.81 frames per second (FPS), while MAGiC-SLAM operates at just over 2 FPS. This performance advantage stems from the use of 3D reconstruction priors in our approach, which predict the 3D structure and thereby reduce the number of iterations required for convergence. As a result, the tracking and mapping processes are significantly accelerated. In contrast, MAGiC-SLAM must estimate the Gaussian primitives from scratch, requiring more iterations to reach convergence and resulting in slower speed.

\section*{Conclusion}

We introduce the first multi-agent monocular dense SLAM system. Leveraging a strong 3D reconstruction prior, our method achieves comparable tracking accuracy to RGB-D SLAM on real-world datasets while significantly reducing processing time. Moreover, our approach can support an arbitrary number of operating agents. However, the performance of the 3D vision prior is relatively sensitive to image resolution, offering an opportunity for future research to develop a more robust 3D foundation model.

\bibliographystyle{IEEEtran}  
\bibliography{Mendeley}       

\vspace{12pt}

\end{document}